\documentclass[journal]{IEEEtran}

\usepackage{booktabs}
\usepackage{tabularx}

\usepackage{pgfplots}
\pgfplotsset{compat=1.17}
\usepackage{tikz}

\usepackage{cite,amsfonts,amsmath,latexsym,amssymb,graphicx,subfigure,cases,booktabs,amsthm,subeqnarray,nomencl,multirow}
\usepackage{pgfplots}
\usepackage{pgfplotstable}
\usepackage{xcolor}
\usepackage{makecell}
\definecolor{myblue}{RGB}{34,31,217}          % for emphasizing text (#5d90c2)
\definecolor{myred}{RGB}{255,66,56}         % for emphasizing text (#cc7b76)
\usepackage[ruled,vlined,lined,ruled,linesnumbered]{algorithm2e}
\usepackage{tikz}
\usepackage[normalem]{ulem}

\usepackage{url}

\usepackage[colorlinks=true,
            citecolor=myred,
            linkcolor=myblue,
            anchorcolor=green,
            urlcolor=green,draft]{hyperref}

\usepackage{bm}
\usepackage{hyperref,bookmark}

\usepackage{textcomp,booktabs}
\usepackage{colortbl}
\definecolor{mygray}{gray}{.9}
\definecolor{mypink}{rgb}{.99,.91,.95}
\definecolor{mycyan}{cmyk}{.3,0,0,0}

\usepackage{color, colortbl}
\newcolumntype{g}{>{\columncolor{Gray}}l}
\newcolumntype{f}{>{\columncolor{Gray}}c}
\definecolor{Gray}{gray}{0.9}

\ifodd 1
\newcommand{\com}[1]{\textbf{\color{red} (COMMENT: #1)}} %comment of the text
\newcommand{\response}[1]{\textbf{\color{green} (RESPONSE: #1)}} %response to comment
\else

\newcommand{\com}[1]{}
\newcommand{\comg}[1]{}
\newcommand{\response}[1]{}
\fi

\newenvironment{proof-sketch}{\noindent{\emph{Sketch of Proof.}}}{\qed\bigskip\\}

\usepackage{nomencl}
\usepackage{mathtools}

\makenomenclature
\RequirePackage{ifthen}
 \renewcommand{\nomgroup}[1]{%
 \ifthenelse{\equal{#1}{p}}{\item[\textbf{Parameters}]}{%
 \ifthenelse{\equal{#1}{v}}{\item[\textbf{Variables}]}{}}}

\IEEEoverridecommandlockouts
 \allowdisplaybreaks 
\begin{document}

\title{\LARGE  Lightweight Multimodal LLM-Enabled Cost-Effective Defect Grading of Power Transmission Equipment}

\author{
	\IEEEauthorblockN{
		Tao~Wang,~\IEEEmembership{Member,~IEEE},  Lipeng~Zhu*,~\IEEEmembership{Senior Member,~IEEE}, Jiayong~Li,~\IEEEmembership{Senior Member,~IEEE},  Feng~Gao,~\IEEEmembership{Member,~IEEE},
		Siwen~Liang,~\IEEEmembership{Member,~IEEE}
		}

	\vspace{-1.1cm}
}

\maketitle

%% Abstract
\begin{abstract}
	
Defect grading of power transmission equipment (DGPTE) is crucial to the stability of electric energy transmission. Although existing machine learning methods exhibit strong capabilities in defect detection, they are plagued by difficulties in integrating expert experience and facing class imbalance in more refined defect grading field. To address this issue, this paper introduces a novel defect grading framework based on multimodal large language model (MLLM). Specifically, this approach maximizes the commercial MLLMs' potential of DGPTE through in-context learning and obtains the state-of-te-art (SOTA) model. By sending a secondary request to this model, a small number of chain of thought-based question-answer pairs (Q\&As) are generated, which effectively reduces the cost of manual annotation. In this way, these high-quality interpretable Q\&As are used to train Qwen3-VL-8B via Low-Rank Adaption-based supervised fine-tuning (SFT). Experimental results on three DGPTE tasks demonstrate that fine-tuning only the language model layer yields the SOTA performance. Furthermore, multi-task joint fine-tuning verifies the feasibility of handling multiple grading tasks within only a single lightweight MLLM.

\end{abstract}


%\vspace{-5mm}
%% Introduction
\section{Introduction}\label{One}

With the rapid development of industrial digitalization, an increasing number of intelligent devices have been deployed in daily operation and maintenance, greatly improving production efficiency. However, industrial equipment often suffers from defects during long-term high-intensity working, severe defects can sometimes lead to unavoidable economic losses, and even casualties. Therefore, it is crucial to conduct periodic effective defect detection after equipment commissioning. \cite{Li11372619, Dong8930292, Shen11052727, LIU10463190, Huang9780551}. As a vital component of the industrial sector, defect detection for power equipment (DDPE) has also received considerable attention \cite{Geng9736589}.

%With the rapid development of intelligent manufacturing, automated defect detection for industrial equipment has become a key technology to ensure production safety. Significant progress has been made in deep learning-based visual inspection methods \cite{Li11372619}\cite{Dong8930292}, weakly supervised anomaly detection techniques \cite{Shen11052727}, vision-language multimodal fusion methods \cite{LIU10463190}, and knowledge-enhanced fault diagnosis systems \cite{LIU10463190}\cite{Huang9780551}, providing diverse technical approaches for industrial defect detection in recent years []. 

In recent decades, many promising achievements have been made in research on DDPE, which has been mainly conducted through two approaches. The first category is traditional physical-property-based defect detection, which identifies defects by analyzing physical signals such as voltage, current, and temperature during equipment operation, combined with empirical thresholds and physical models \cite{Huang10387819, Meitei9355190, CHAOUCHE2026112722}. However, such methods rely on manual feature design and have weak adaptability to complex working conditions. The second category is data-driven detection methods \cite{Lin180341, Wang814667, Hu11373592}. Based on massive monitoring data and algorithms such as deep learning, these methods automatically mine potential information of defects, showing stronger detection ability. In fact, they have gradually become the mainstream research direction, a considerable number of review articles on AI-based DDPG have been completed for different power equipment \cite {Wu10818631, Aydin2025ARO, Wang10134581}.

%As an important component for maintaining social operation, the power industry has also attracted extensive attention to power equipment defect detection (PEDD) since the late 20th century \cite{Lin180341}\cite{Wang814667}. Reference \cite{Zaben10416843} reviews machine learning methods for microgrid fault detection, including support vector machines (SVM), DT, and neural networks (NN), which can handle complex nonlinear relationships. Liu et al. \cite{Su9136904} designs a complementary attention network (CAN) for defect detection in electroluminescence (EL) images of solar cells, which suppresses background noise in defect detection through channel and spatial attention mechanisms. Reference \cite{Gonzalez11014155} proposes a hybrid method combining physical models and data-driven approaches, which uses one-class SVM (OCSVM) trained on healthy data to alleviate the scarcity of fault data. However, these traditional algorithms can only rely on single-modal data such as sensor or image data. 

Nevertheless, data-driven methods often require large-scale training data to achieve advanced performance. Due to the large span of transmission lines and the need to cross complex terrain, power transmission equipment (PTE) is usually deployed in high-altitude, sparsely populated mountainous areas, which makes the information collection very difficult, thus led to slow progress in the specific field of intelligent defect detection of power transmission equipment (DDPTE) in the early stage. Benefiting from advanced transmission line inspection technology via UAV \cite{Huang9991049}, the problem of data acquisition for DDPTE has been solved, and research on accomplishing DDPTE using computer vision algorithms based on UAV images has begun to emerge \cite{Huang9991049, Zhang11407486, YIN2024100814, Cao10663474, He11028102}. It is worth mentioning that the advanced performance demonstrated in these studies is based on a large number of images, good shooting quality, and diverse equipment angles. However, such a dataset requires the accumulation of a large number of UAV inspections over a long period of time. Some studies aim to address this limitation through image data augmentation \cite{Di11216605}, but this is highly dependent on the quality of sample generation. Moreover, data-driven methods still lack interpretability; humans can only enable models to indirectly understand human intentions through a great deal of manual annotation, rather than directly integrating expert experience.

% Yin et al. \cite{YIN2024100814} designs an intelligent detection method based on data mining, which combines UAV aerial photography and clustering analysis to realize automatic defect classification. Cao et al. \cite{Cao10663474} proposes a lightweight CACS-YOLO model with channel shuffle and coordinate attention, reducing model parameters by 60\% while maintaining accuracy. To address the lack of high-quality training samples, Di et al. \cite{Di11216605} proposes a few-shot learning (FSL) method based on image augmentation and reconstruction. He et al. \cite{He11028102} presents a weakly supervised contrastive learning pre-training method for pixel-level defect localization with only image-level labels, significantly reducing annotation costs. However, the quality of generated samples and the construction of positive and negative samples greatly affect learning performance, which restricts the deep application of intelligent detection in practical engineering.

Recently, MLLM, which inherently possesses human-like visual reasoning ability \cite{Han11272128} and generalization under few-shot or zero-shot setting \cite{GUO2025126835}, has shown great potential in this field \cite{Li11262214, Zhou11052760}. For example, Zhao et al. \cite{Zhao10879360} proposes CMKR-PBDM, which constructs a small number of image-text dataset and a bolt knowledge graph to achieve interpretable detection via knowledge-enhanced  reasoning. Li et al. \cite{Zhang11076154} proposes VLF-DETR, which completes DDPTE tasks by fine-tuning FLAVA through three steps: line anomaly knowledge injection, image fusion, and edge feature extraction. These studies have utilized the capabilities of MLLM, which computer vision algorithms do not possess, to conduct pioneering work in the field of DDPTE, but the utilization of MLLM remains limited.

%MLLM inherently possesses advanced human-like visual reasoning ability \cite{Han11272128} and strong generalization under few-shot and  settings \cite{GUO2025126835}, and have shown promising performance in industrial defect detection \cite{Li11262214}. Zhou et al. \cite{Zhou11052760} proposes the T2MFDF framework, which integrates time series and text descriptions and uses LLM to achieve cross-modal alignment and knowledge fusion, significantly improving bearing fault diagnosis accuracy. For transmission line inspection, Zhao et al. \cite{Zhao10879360} proposes CMKR-PBDM, which constructs an image-text dataset and a bolt knowledge graph to address limited information and insufficient global reasoning ability, achieving interpretable detection via knowledge-enhanced LLM reasoning. Zhang et al. \cite{Zhang11076154} proposes VLF-DETR, which fuses vision-language and high-frequency features for transmission line defect detection. Although MLLM can significantly improve the intelligence of equipment defect detection by fusing multi-source information. However, fine-tuning MLLM requires massive high-quality image-text pairs, API calls incur high costs, and using cloud-based MLLM faces data leakage risks.

To further explore the MLLM's capacity in this field, we first select a more challenging task than DDPTE—Defect Grading of Power Transmission Equipment (DGPTE), which is a more refined defect detection, and aims to judge the defect level after the occurrence of defect. Due to the heavy reliance on expert knowledge, electrical engineer integrates several single-target visual models and logic codes into an orderly pipeline to solve a specific DGPTE task. This has brought a series of limitations, such as the need for more samples to train multiple visual models and error accumulation in each step. For this reason, the accuracy of automated operations for DGPTE is far lower than that of manual work. What's more, this task often faces a serious long-tail sample problem, so relevant research remains limited, even on a broader scale \cite{Fei10082595, electronics14153101, Zheng2024EnhancingPE}. Based on this situation, this paper proposes a novel MLLM-enabled strategy to efficiently accomplish DGPTE tasks, which tentatively solves all the limitations mentioned in the above literature review, enabling the accuracy of MLLM to slightly exceed the human grading level while endowing the grading process with interpretability. The main contributions of this paper are as follows:

\begin{enumerate}
	\item This work proposes a novel GDPTE solution to address the challenges of long-tail problem and difficulty in fusing human knowledge. Based on an effective fine-tuning strategy, a light-weight MLLM can achieve human-like defect grading in a cost-effective manner. To the best of the authors' knowledge, this is the first work to apply MLLM to accomplish this task.

	\item This work designs efficient prompts to obtain the state-of-the-art (SOTA) model from commercial MLLMs. By sending a secondary request to the model, a small number of Chain of Thought-based Q\&A for subsequent fine-tuning are generated, this process significantly reduces the amount of manual annotation acquired.
	
	\item This work fine-tunes Qwen3-VL-8B on three PTEDG tasks under the few-shot setting. The fine-tuned MLLM significantly outperforms commercial MLLMs and achieves SOTA performance when only fine-tuning the LLM layer. Meanwhile, multi-task joint fine-tuning verifies the feasibility of integrating multiple defect grading tasks into a single-model pipeline.

%	Based on two carefully designed prompts, first evaluating multiple commercial MLLMs to select the SOTA model, and then utilizing it to generate a small number of high-quality Q\&A with CoT reasoning processes to support fine-tuning.
%	
%	\item We perform LoRA-based SFT on the open-source restricted MLLM—Qwen3-VL-8B—for the three selected PTEDG tasks under the FSL setting. Experiments demonstrate that the proposed fine-tuning framework achieves a substantial improvement, and the most significant performance is obtained by only fine-tuning LLM layer in it. Meanwhile, we perform joint fine-tuning of all chosen tasks and observe no significant accuracy decline. This finding verifies the feasibility of simultaneously meeting the requirements of only a small amount of fine-tuning data, ensuring local safe calling, and unifying multiple tasks into a single-model pipeline.

\end{enumerate}

\begin{table}[htbp]
	\centering
	\caption{Commonly Used Symbols}
	\begin{tabularx}{3.5in}{@{}lX@{}}
		\toprule 
		\textbf{Acronym} & \textbf{Description} \\ 
		\midrule 
		
		DGPTE            & Defect grading of power transmission equipment \\
		
		MLLM             & Multimodal Large Language Model \\
		SFT              & Supervised Fine-tuning \\
		LoRA             & Low-Rank Adaptation \\

		ICL			     & In-context learning \\
		FSL              & Few-shot learning \\
		DT               & Decision Tree \\
		CoT              & Chain of thought \\
		Q\&A             & Question-answer pairs \\

		VE               & Visual Encoder layer \\
		MMA              & Multimodal Alignment layer \\
		LLM              & Language Language Model layer \\
		
		MF1              & Mecro-F1-Score \\
		ACC              & Accuracy \\

		Num Pre          & Number of predicts \\
		Num Ctx          & Number of context tokens \\	
		SOTA	         & State-of-the-art           \\	
		API	             & Application programming interface           \\

		\midrule 
		\textbf{Parameter} & \textbf{Description} \\ 
		\midrule 
		$\mathcal{R}^{dg}$     & Prompt for PTEDG \\
		$\mathcal{R}^{qa}$ & Prompt for  Q\&A generation\\ 
		$m^*$ & SOTA MLLM for PTEDG\\ 
		$\mathcal{M}$     & Commercial MLLMs Set \\
		$\hat{\mathcal{A}}^{dg}$     & Predicted  DT-based CoT result of PTEDG \\
		$\mathcal{A}^{dg}$ & True DT-based CoT result of PTEDG\\ 
		$\mathcal{C}$ & Fine-tuning module set \\ 
		\bottomrule 
		
	\end{tabularx}
	\label{tab:acronyms}
	\vspace{-5mm} 
\end{table}

% TASK DESCRIPTION AND PROPOSED FRAMEWORK
\section{Description of Selected PTEDG Tasks}\label{Two}

In this paper, we select three DGPTE tasks of varying difficulty to verify the effectiveness of the proposed framework.

%, as summarized in Table \ref{Task Definition for Defect Grading}.
%\begin{table}[h]
%	\centering
%	\caption{Task Definition for Defect Grading}
%	\begin{tabular}{@{}lll@{}}
%		\toprule
%		\textbf{Task} & \textbf{Name} & \textbf{Grade} \\
%		\midrule
%		Task 1 & Tension Clamp – Crimping & Else, Kind \\
%		Task 2 & Guying Fitting – Corrosion & Else, Kind, Major \\
%		Task 3 & Tangent Tower – Bird's Nest & Else, Kind, Major, Urgent \\
%		\bottomrule
%	\end{tabular}
%	\label{Task Definition for Defect Grading}
%	\vspace{-2mm} 
%\end{table}

\begin{figure}[t]
	\centering
	\subfigure[Task 1: Tension Clamp – Crimping]{
		\includegraphics[width=3.5in]{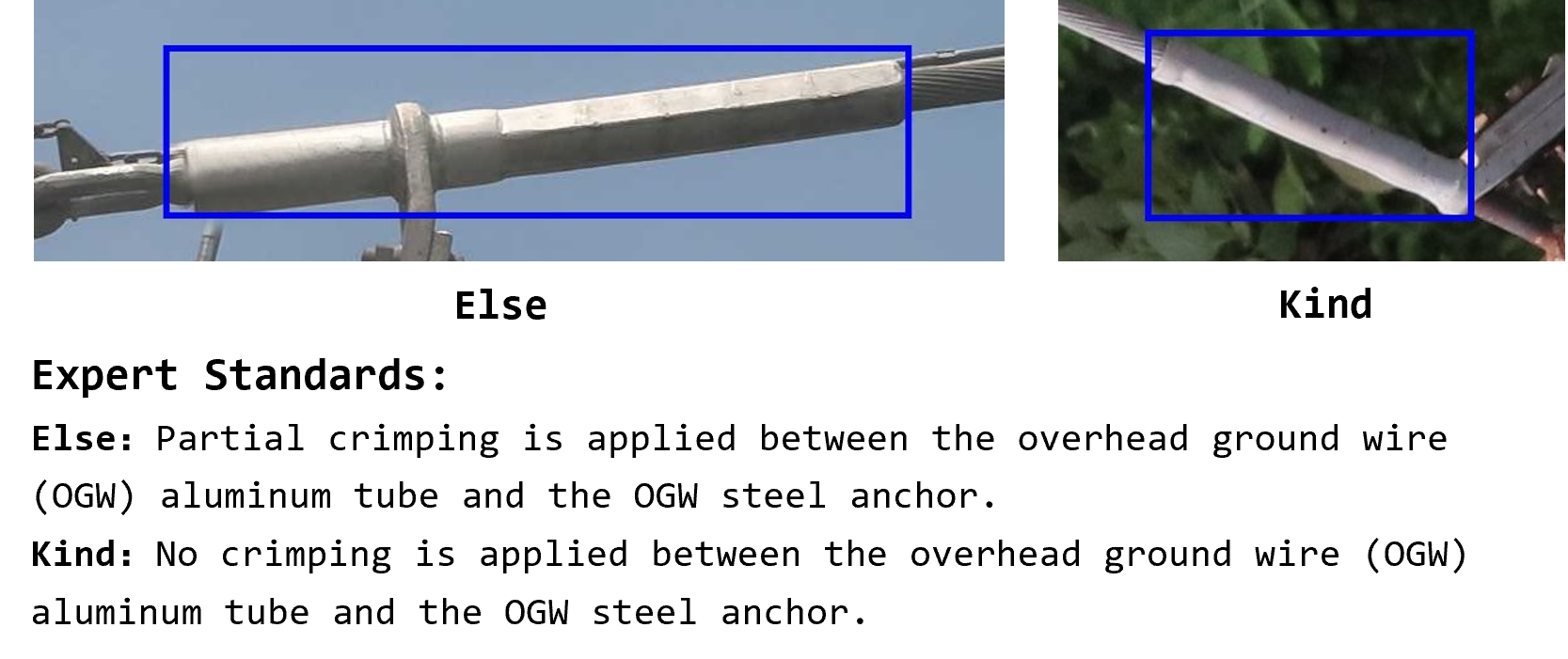} 
		\label{Tension Clamp – Crimping}
	}

	\subfigure[Task 2: Guying Fitting – Corrosion]{
		\includegraphics[width=3.5in]{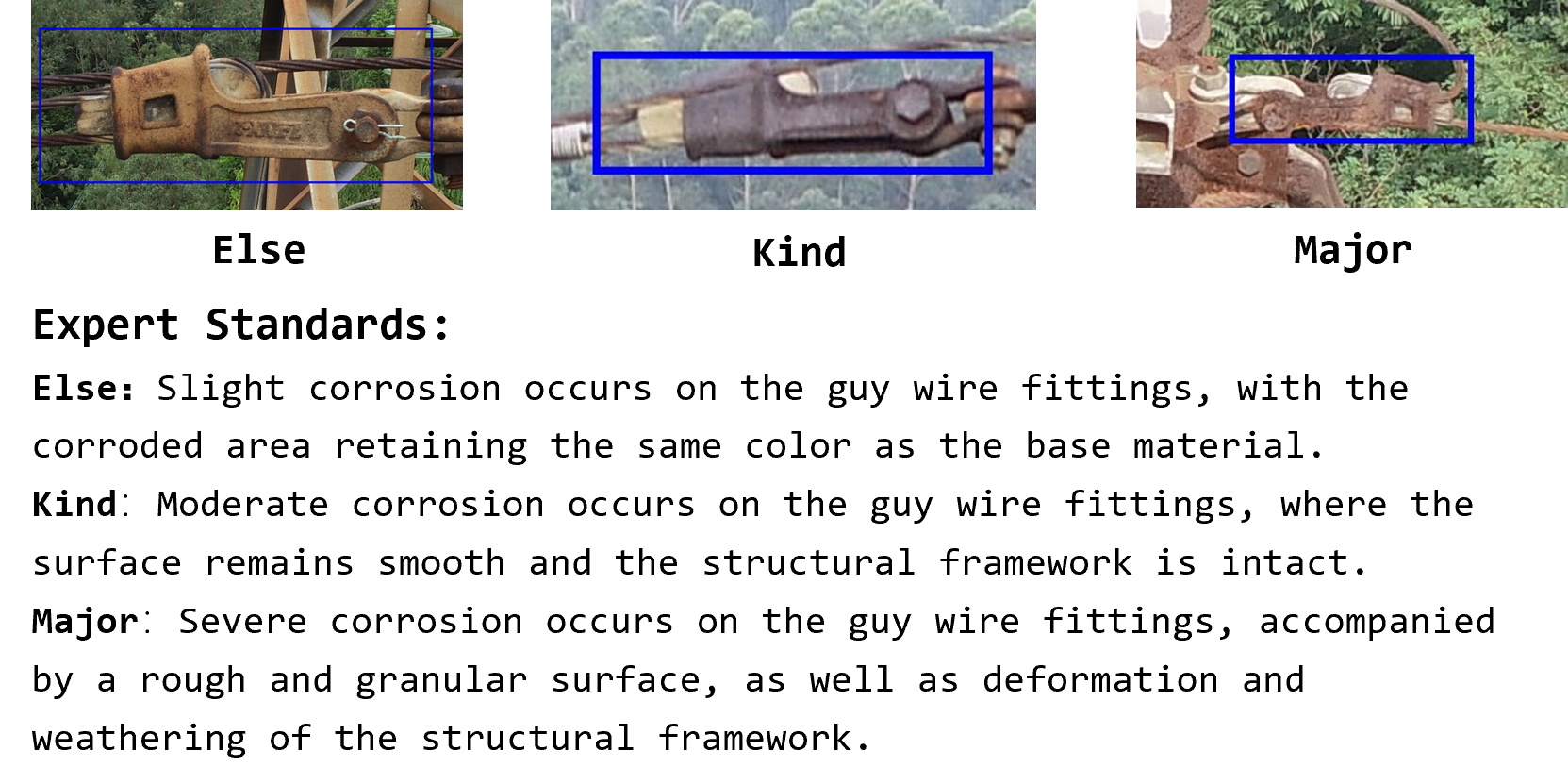}
		\label{Guying Fitting – Corrosion}
	}

	\subfigure[Task 3: Tangent Tower – Bird's Nest]{
		\includegraphics[width=3.5in]{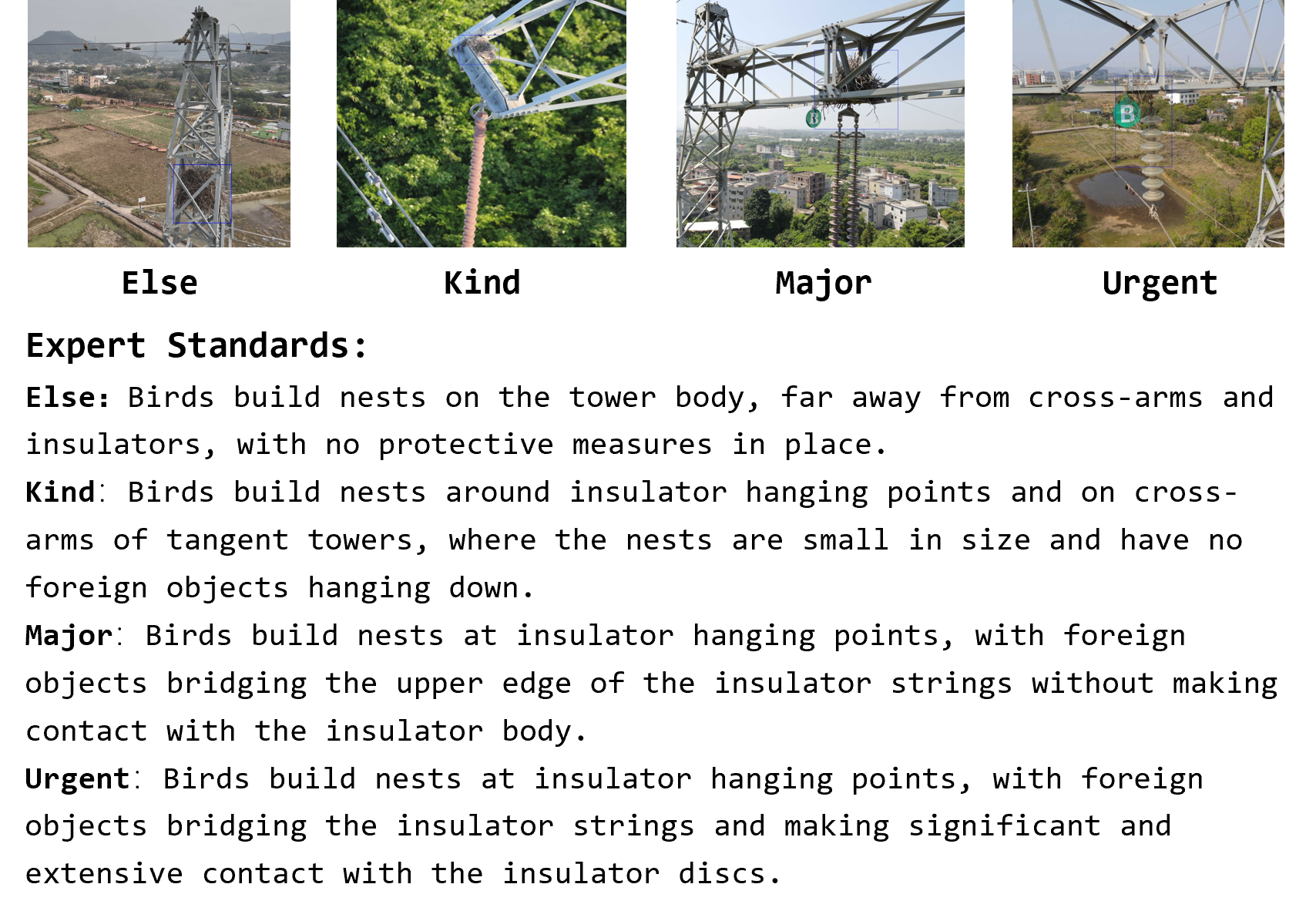}
		\label{Tangent Tower – Bird's Nest}
	}
	\vspace{-2mm}
	\caption{The three PTEDG tasks used in the experiments of this paper}
	\label{fig_1}
\end{figure}

Task 1 is dedicated to identifying the crimping status of tension clamps, which is consisting of two categories: Else and Kind. Crimping denotes a mechanical connection technique that relies on high pressure to bond clamps and conductors permanently and rigidly. Two representative examples of tension clamps, together with the expert-led grading workflow, are illustrated in Fig. \ref{Tension Clamp – Crimping}. This is a binary classification task that requires the model to possess component localization and shape recognition capabilities.

For Task 2, the objective is to assess the corrosion degree of tension hardware by analyzing its surface color variations and structural damage characteristics. Framed as a three-class classification task, it encompasses three defect levels: Else, Kind, and Major. Corresponding task instances and the expert standards are presented in Fig. \ref{Guying Fitting – Corrosion}. This is a three-class classification task that requires component localization, color identification, and area estimation capabilities.

Task 3 focuses on evaluating the potential impact of bird nests on insulators, with the assessment criteria incorporating two key factors: the nest position on tangent towers and the actual state of the nests themselves. This task is designed as a four-class classification task, covering four hierarchical defect levels: Else, Kind, Major, and Urgent. Comprehensive details regarding the task definition and sample cases are provided in Fig. \ref{Tangent Tower – Bird's Nest}. This is a four-class classification task that requires multiple capabilities such as 3D localization.

%\vspace{-2mm}

% Methodology
\section{Proposed Methodology}\label{Three}

\begin{figure*}[ht]
	\centering
	\includegraphics[width=7in]{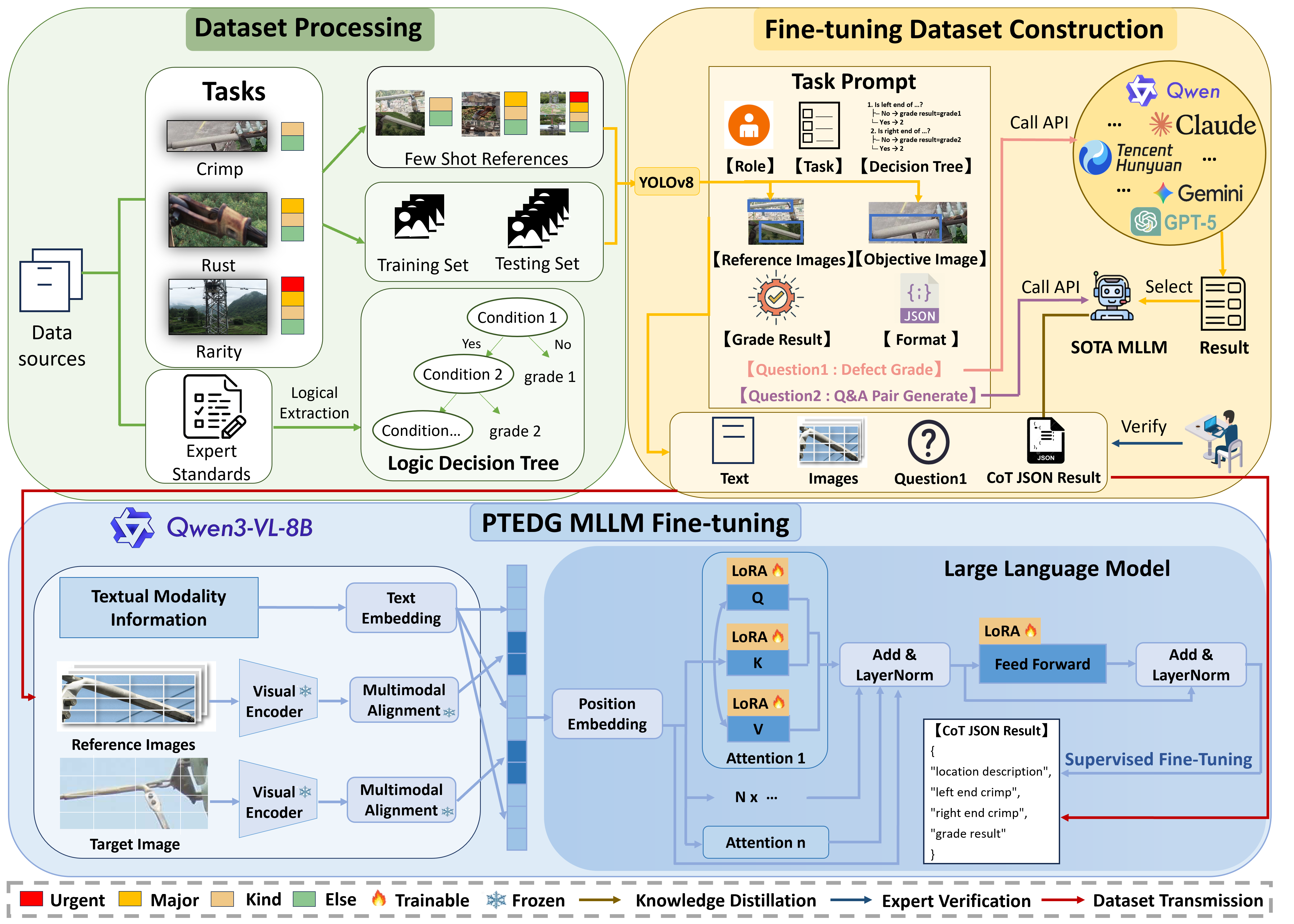}
	\vspace{-2mm}
	\caption{The framework of proposed fine-tuning strategy}
	\label{proposed fine-tuning strategy}
	\vspace{-2mm}
\end{figure*}

To better leverage MLLM to handle the above DGPTE tasks, the proposed defect grading framework elaborately designs a complete pipeline, the detailed workflow is illustrated in Fig. \ref{proposed fine-tuning strategy}. In the Dataset Processing module, we uniformly process multimodal data from three tasks. Subsequently, we query commercial MLLMs to select the SOTA defect grading model, which is then used to generate high-quality Q\&As for SFT in the Fine-tuning Dataset Construction module. Finally, we utilize this Q\&As to fine-tune different components of Qwen3-VL-8B in the DGPTE MLLM Fint-tuning module. The more details will be mentioned in the following subsections.

%\vspace{-2mm}

\subsection{Dataset Processing Module}

This subsection is illustrated as Dataset Processing in Fig. \ref{proposed fine-tuning strategy}. For the image data, we perform the image resizing operation to ensure that MLLM can adapt to a fixed size and achieve the desired performance during fine-tuning. Specifically, we first check if either side of the image has a pixel dimension greater than or equal to 1280. If not, the original image size is retained; if yes, the longer side is scaled to 1280 pixels, and the shorter side is scaled proportionally. The detailed specifics are presented in Equation \eqref{image_resize}:

\begin{equation}
	\vspace{-2mm}
	\small
	\label{image_resize}
	(W', H') =
	\begin{cases}
		(W, H) & \max(W, H) < 1280, \\
		\left(1280, \frac{H \times 1280}{W}\right) & W = \max(W, H) \land W \geq 1280, \\
		\left(\frac{W \times 1280}{H}, 1280\right) & H = \max(W, H) \land H \geq 1280.
	\end{cases}
\end{equation}
\vspace{2mm}
where $W$, $H$ are width and height of the original image, respectively. $W'$, $H'$ are width and height of the resized image, respectively.  $\max(W, H)$ is the maximum value of image width and height.

In view of the long-tail sample problem general existing in the real world, we complete this work under the few-shot learning (FSL) setting \cite{Perez10.5555}. Specifically, the dataset is partitioned into the training and testing subsets following a strict stratified sampling strategy: only 30 images are selected for each defect grading within every task, with the remaining images assigned to the test subset. Therefore, 60 images of Task 1, 90 images of Task 2, and 120 images of Task 3 are utilized for MLLM fine-tuning. Besides, 1 typical image of each grade is selected and incorporated into the prompt as main component of in-context learning (ICL) \cite{dong-etal-2024-survey} to help MLLM grasping task-specific human knowledge.

For the expert standards, they are extracted and enhanced logically to Decision Tree-based Chain-of-Thought (DT-based CoT). As a classical machine learning model, DT differs from deep learning in that it offers interpretability. Notably, a key advantage of MLLM lies in its ability to explicitly generate reasoning processes by CoT \cite{10.5555/3600270.3602070} while delivering end-to-end task outputs. Therefore, these two methods complement each other and DT-based CoT can further improve the logical rigor and reliability of the MLLM's output. Table \ref{Decision Tree-based Chain of Thought} illustrates the DT-based CoT process for Task 3, which requires human experts to perform logical decomposition.

\begin{table}[htbp]
	\centering
	\caption{Decision Tree-based Chain of Thought with Task 3}
	\label{Decision Tree-based Chain of Thought}
	\begin{tabular}{@{}>{\raggedright\arraybackslash}p{\linewidth}@{}}
		\noalign{\hrule height 1pt}
		\vspace{0.1pt}\textbf{Execute the checks in steps 1$\to$4}. If a \textbf{No} or \textbf{Not} is returned, the grading outcome shall be determined directly, and the subsequent check items are exempted from further evaluation.\vspace{3pt} \\
		\noalign{\hrule height 0.4pt}
		\vspace{1pt} \textbf{1. Having insulator:} Does the image contain at least one insulator string? \\
		\quad $\boldsymbol{\vdash}$ Not Exists $\to$ grade result is "Else" \\
		\quad $\boldsymbol{\vdash}$ Exists $\to$ 2 \\[4pt]
		\textbf{2. Nest location:} Is the bird's nest located on the crossarm, away from the main body of the vertical tangent tower or concrete pole? \\
		\quad $\boldsymbol{\vdash}$ No $\to$ grade result is "Else" \\
		\quad $\boldsymbol{\vdash}$ Yes $\to$ 3 \\[4pt]
		\textbf{3. Drooping:} Do the straw, branches, or other materials hang down? \\
		\quad $\boldsymbol{\vdash}$ No $\to$ 4A \\
		\quad $\boldsymbol{\vdash}$ Yes $\to$ 4B \\[4pt]
		\textbf{4A. Nest Structure:} Is the bird's nest loose or of large volume? \\
		\quad $\boldsymbol{\vdash}$ No $\to$ grade result is "Kind" \\
		\quad $\boldsymbol{\vdash}$ Yes $\to$ grade result is "Major" \\[4pt]
		\textbf{4B. Drooping Length:} Are the drooping materials long enough to clearly exceed the length of the metal hangers connecting the insulators and tangent tower, and have they wrapped around the insulator? \\
		\quad $\boldsymbol{\vdash}$ No $\to$ grade result is "Major" \\
		\quad $\boldsymbol{\vdash}$ Yes $\to$ grade result is "Urgent"\vspace{3pt} \\
		\noalign{\hrule height 1pt}
	\end{tabular}
	\vspace{-5mm} 
\end{table}

\subsection{Fine-tuning Dataset Construction}

This subsection introduces the details of two crucial prompts in this framework and the process of calling the commercial MLLMs to generate Q\&As for subsequent model fine-tuning, as shown in the Fine-tuning Dataset Construction module of Fig. \ref{proposed fine-tuning strategy}. It concludes three steps below:

\begin{itemize}
	\item \textit{Step 1:} Input all images to YOLOv8 for object detection, where the positions of objective equipment need to be labeled in the images. It provides precise guidance for subsequent defect grading.
	\item \textit{Step 2:} Design defect grading prompt $\mathcal{R}^{dg}$, feed it into the commercial MLLMs and select the SOTA MLLM for the target tasks:
	
	\begin{equation}
		\label{select_sota_single}
		m^* = \underset{m \in \mathcal{M}}{\operatorname{argmax}} \; ACC_{m({\mathcal{R}_{dg}})}
	\end{equation}
	where $m^*$ and $\mathcal{M}$ denote the SOTA MLLM and the candidate MLLMs, respectively, and $ACC_{m(\mathcal{R}^{dg})}$ represents the MLLMs' grading accuracy when input $\mathcal{R}^{dg}$.

	\item \textit{Step 3:} Develop Q\&As generating prompt  $\mathcal{R}^{qa}$ to call the SOTA MLLM again and generate a small number of Q\&As. After human experts' verification, the final Q\&As for fine-tuning Qwen3-VL-8B are obtained. The details is as shown in Equation \eqref{qa_generation}:

	\begin{equation}
		\label{qa_generation}
		\mathcal{Q}\&\mathcal{A} = m^*\left(\mathcal{R}^{qa}\right)
	\end{equation}
	
\end{itemize}

\begin{figure}[htbp]
	\centering
	\includegraphics[width=3.5in]{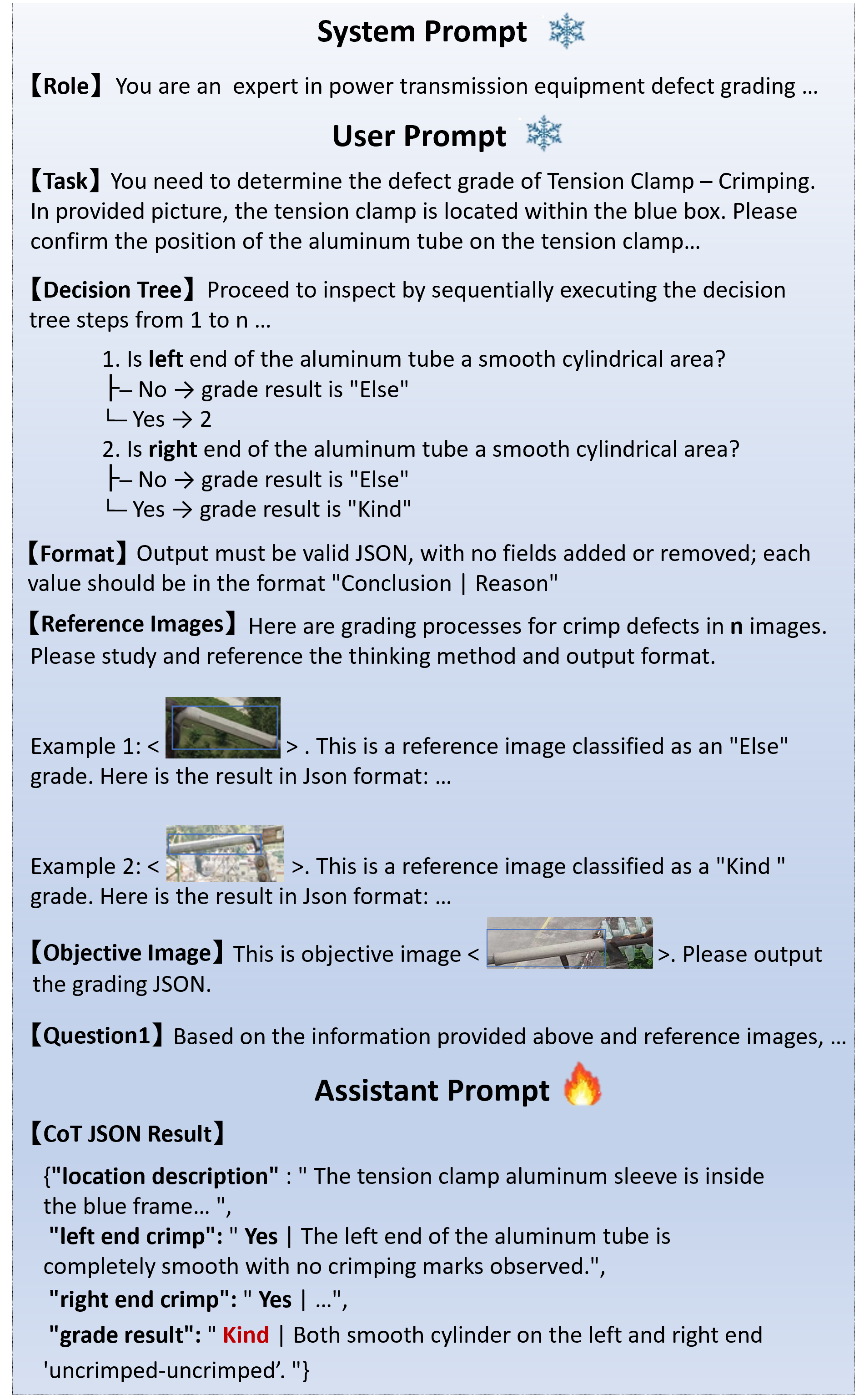}
	\caption{An example of Q\&A for MLLM fine-tuning in Task 1}
	\vspace{-5mm}
	\label{The details of prompt}
\end{figure}

In Fig. \ref{The details of prompt}, the System Prompt (SP) block and User Prompt (UP) block show an example of the composition of $\mathcal{R}_{dg}$ under Task 1, the composition of $\mathcal{R}_{qa}$ is slightly different but largely consistent. Specifically, we define $\mathcal{R}_R$ as the part of Role, which is equivalent to SP and where we specify that MLLM needs to slove a DGPTE problem. In UP, Task $\mathcal{R}_T$ defines the MLLMs' target of DGPTE at this time and receives some prior information. Decision Tree $\mathcal{R}_{DT}$ incorporates DT-based grading logic, which has been introduced in subsection A. Reference Images $\mathcal{R}_{RI}$ denote the reference images and their DT-based CoT process as the important grading references. Objective Image $\mathcal{R}_{OI}$ is the current target grading image. Format $\mathcal{R}_F$ specifies an output structure with the standard JSON format, which requires MLLM to generate step-by-step intermediate and final results following $\mathcal{R}_{DT}$. 

It is worth noting that Grade Result $\mathcal{R}_{GR}$ represents the pure grading results annotated by experts, without any intermediate steps. $\mathcal{R}_{GR}$ is utilized in $\mathcal{R}_{qa}$ to guide the SOTA MLLM in deriving the reasoning process based on the true answer. Finally, Question1 ($\mathcal{R}_{\mathcal{Q}_1}$) and Question2 ($\mathcal{R}_{\mathcal{Q}_2}$) serve as the question component of $\mathcal{R}_{dg}$ and $\mathcal{R}_{qa}$, respectively, with details provided as follows:

\textit{Q1: Based on the information provided above and the reference images, please determine the defect grade of the objective image.}

\textit{Q2: Based on the information provided above, together with the objective image and its real defect grading result, you are required to generate the CoT reasoning process and output it in the JSON format specified in the provided examples.}

Therefore, $\mathcal{R}^{dg}$ and $\mathcal{R}^{qa}$ are expressed as the following Equations \eqref{RDG} and Equations \eqref{RQA}, respectively:

\begin{subequations} \label{loss-terms-main} 
	\begin{align}
		&\smaller \mathcal{R}^{dg} = \mathcal{R}_R + \mathcal{R}_T + \mathcal{R}_{DT} + \mathcal{R}_F + \mathcal{R}_{RI} + \mathcal{R}_{TI} + \mathcal{R}_{\mathcal{Q}_1}, \label{RDG} \\
		&\smaller \mathcal{R}^{qa} = \mathcal{R}_R + \mathcal{R}_T + \mathcal{R}_{DT} + \mathcal{R}_F + \mathcal{R}_{RI} + \mathcal{R}_{TI} + \mathcal{R}_{GR} + \mathcal{R}_{Q_2} \label{RQA}
	\end{align}
\end{subequations}

After calling the SOTA MLLM and generating Q\&A for each image in the training set, human experts are required to inspect the reasoning process to ensure full correctness. Since the size of this dataset is relatively small, this step minimizes the manual annotation cost as much as possible.

Assistant Prompt, i.e., the CoT JSON Result in Figure 3, presents an example of CoT-based answers for Task 1, which analyzes several key factors that influence the final grading result. Finally, we compose Q\&As from Defect Grading Prompt  $\mathcal{R}^{dg}$ posed for each training sample and the corresponding result $\mathcal{A}^{dg}$.

\subsection{DGPTE MLLM Fine-tuning Module}
After obtaining the Q\&A pairs available for fine-tuning, we introduce the architecture of MLLM and LoRA-based supervised fine-tuning process, as illustrated in Figure 1: DGPTE MLLM Fine-tuning module. Herein, to facilitate the understanding of the formulas, we describe the fine-tuning process from the perspective of a single Q\&A.

\textit{1) MLLM Architecture:} The popular architecture is composed of three components, namely visual encoder (VE) layer, multimodal alignment (MMA) layer and large language model (LLM) layer. When a multimodal question $\mathcal{R}^{dg}$ is input into MLLM, the textual modal information is encoded by the pre-trained text embedding layer (TE). And the visual modal information, including reference images and objective image, is sequentially encoded by the VE and MMA in Euation \eqref{text_visual_encoding}:

\begin{align}
	&\boldsymbol{h}_i =
	\begin{cases}
		\boldsymbol{h}_i^t & \text{if } \boldsymbol{x}_i \text{ is textual information} \\
		\boldsymbol{h}_i^v & \text{if } \boldsymbol{x}_i \text{ is visual information}
	\end{cases}, \label{text_visual_encoding} \\
	&\boldsymbol{h}_i^t = E_t(\boldsymbol{x}_i), \quad \boldsymbol{h}_i^v = E_m(E_v(\boldsymbol{x}_i)) \notag
\end{align}
where $\boldsymbol{x}_i$ represents the $i$-th input token, $\boldsymbol{h}_i^t \in \mathbb{R}^{L_t \times d_{llm}}$, $\boldsymbol{h}_i^v \in \mathbb{R}^{L_v \times d_{llm}}$ are the textual and visual information, respectively. Here, $E_t(\cdot)$ denotes the TE function, $E_v(\cdot)$ is the VE function, and $E_m(\cdot)$ represents the MMA function.

Subsequently, the all embedding of input information is concatenated according to their relative positions, with positional encoding added, and then fed into the LLM layer, which autoregressively generates the answer $\hat{\mathcal{A}}^{dg}$ in the MLLM's inference stage, as shown in Equation \eqref{concat_inference}.

\begin{align}
	&\boldsymbol{H}_{concat} = \left[\boldsymbol{h}_1; \boldsymbol{h}_2; \dots; \boldsymbol{h}_m\right] + \boldsymbol{P}_{1:m} \label{concat_inference} \\
	&\hat{\mathcal{A}}^{dg} = f_{LLM}(\boldsymbol{H}_{concat}) \notag
	\end{align}
where $\boldsymbol{H}_{concat} \in \mathbb{R}^{L \times d_{llm}}$ denotes the concatenated vector of all multimodal information, $\boldsymbol{P}$ represents the positional encoding function, and $f_{LLM}$ is the LLM function.

\textit{2) LoRA-based SFT Process:} Low-Rank Adaption (LoRA) \cite{hu2022lora} is an efficient fine-tuning technique, which freezes the original weights $\boldsymbol{W}$ of MLLM during fine-tuning and decomposes the weight update matrix $\Delta \boldsymbol{W}$ into two low-rank matrices $\boldsymbol{B}$ and $\boldsymbol{D}$, and trains only these two matrices. After that, it merges $\boldsymbol{B}$ and $\boldsymbol{D}$ to recover $\Delta \boldsymbol{W}$, and then integrates $\Delta \boldsymbol{W}$ into the original weights $\boldsymbol{W}$ to form the updated weights $\boldsymbol{W}'$, as shown in the following equation \eqref{lora}:

\begin{equation}\label{lora}
	\boldsymbol{W}' = \boldsymbol{W} + \Delta \boldsymbol{W} = \boldsymbol{W} + \boldsymbol{B}\boldsymbol{D}
\end{equation}where $\boldsymbol{B} \in \mathbb{R}^{d \times r}$ and $\boldsymbol{D} \in \mathbb{R}^{r \times k}$, with a low rank $r \ll \min(d, k)$.

This paper adjusts the parameter of Qwen3-VL-8B via LoRA-based SFT, where the parameters of any one or more modules among VE, MMA, and LLM layers can be fine-tuned. The formula for its loss function $\mathcal{L}$ is shown as follows:

\begin{equation}
	\label{Loss function}
	\mathcal{L}_{SFT}(\theta) = -\sum_{j=1}^T \log P_\theta\left(\mathcal{\hat{A}}_j^{dg} \mid \mathcal{R}^{dg}, \mathcal{A}_{<j}^{dg}\right)
\end{equation}
where $\mathcal{A}_{<j}^{dg}$ and $\mathcal{\hat{A}}_j^{dg}$ denote the first $j-1$ ground-truth tokens and the $j$-th predicted token of the defect grading result, respectively. $\mathcal{L}$ needs that each subsequent token is sampled based on all preceding ground-truth tokens in single sample, following the probability distribution $P_\theta$. The ultimate objective of SFT is to update the model parameters $\theta$ given $\mathcal{R}^{dg}$ and $\mathcal{A}_{<j}^{dg}$, such that LLM can sample the autoregressively output token $\mathcal{\hat{A}}_j^{dg}$ with the maximum probability, which matches the ground-truth token $\mathcal{A}_{j}^{dg}$. 

%Thus, the consistency between the predicted $\hat{\mathcal{A}}^{dg}$ and the ground-truth $\mathcal{A}^{dg}$ can be maximized during inference.

Based on Equation \eqref{Loss function}, we can derive the formulas for updating the parameters of matrices $\boldsymbol{B}$ and $\boldsymbol{D}$ using gradient descent as follows:

\begin{subequations} \label{eq:param_update_components-main} 
	\begin{align}
		&\boldsymbol{B} \leftarrow \boldsymbol{B} - \eta \nabla_{\boldsymbol{B}} \mathcal{L}_{SFT}(\theta_\mathcal{C}), \label{eq:param_update-B} \\
		&\boldsymbol{D} \leftarrow \boldsymbol{D} - \eta \nabla_{\boldsymbol{D}} \mathcal{L}_{SFT}(\theta_\mathcal{C}), \label{eq:param_update-D} \\
		&\mathcal{C} \subseteq \left\{ \text{VE}, \text{MMA}, \text{LLM} \right\} \notag  % \notag 取消编号，保留标签（可选）
	\end{align}
\end{subequations}
here, $\eta$ denotes the learning rate, $\nabla_{\boldsymbol{B}}$ and $\nabla_{\boldsymbol{D}}$ represent the gradients with respect to matrices $\boldsymbol{B}$ and $\boldsymbol{D}$, respectively, and $\mathcal{C}$ denotes the set of modules with fine-tunable parameters.

\section{EXPERIMENTS AND ANALYSIS}\label{Four}
\subsection{Experimental Settings and Evaluation Metrics}

This paper employs Qwen3-VL-8B \cite{li2026qwen3vlembeddingqwen3vlrerankerunifiedframework} as the base MLLM for fine-tuning, LLaMA Factory\footnote{Available: \url{https://github.com/hiyouga/LlamaFactory}} and Ollama\footnote{Available: \url{https://github.com/ollama/ollama}} are adopted as the fine-tuning framework and deployment framework, respectively. All experiments are conducted on a machine equipped with 8*NVIDIA-A100-80G GPUs, with a total video memory of 640 GB. Meanwhile, nine commercial MLLMs are employed for the defect grading test by invoking official APIs, and the SOTA model is further queried to generate Q\&As, including six models in Qwen and GPT series and three model from Gemini, Hunyuan, and Claude.

The basic experimental parameters are shown in Table \ref{Fine-tuning and Inference Parameters}. In the fine-tuning stage, batch size, and epoch are dynamically adjusted to ensure better fine-tuning performance under 640 GB of available video memory. LoRA rank and LoRA alpha represent the scale of fine-tuning parameters and the fine-tuning intensity, which are set to 32 and 128, respectively. To ensure MLLM produces corresponding outputs while balancing accuracy and computational resource consumption, the maximum image resolution is set to 1280×1280, and the Cutoff length is set to 12,280. In the inference stage, temperature is set to 0 to ensure the consistency of results. At this time,  TOP P and TOP K are invalid. Num Ctx refers to the context length supported by Qwen3-VL-8B, which is set to the default maximum supported length of 256k. Num Pre occupies fixed video memory for reserved KV cache when the model is launched, and it is set to be consistent with Cutoff ength.

\begin{table}[!t]
	\centering
	\caption{Fine-tuning and Inference Parameters}
	\label{Fine-tuning and Inference Parameters}
	\renewcommand{\arraystretch}{1.2} 
	\resizebox{\linewidth}{!}{
		\begin{tabular}{l l c}
			\hline
			\textbf{Stage} & \textbf{Parameter} & \textbf{Range / Value} \\
			\hline
			\multirow{10}{*}{\makecell[l]{Fine-tuning}} 
			& Fine-tuning Model & Qwen3-VL-8B \\
			& Fine-tuning Type & SFT \\
			& Fine-tuning Module & \{VE, MMA, LLM\} \\
			& Fine-tuning Method & LoRA \\
			& LoRA Rank & 32 \\
			& LoRA Alpha & 128 \\
			& Batch Size & \{1, 2\} \\
			& Cutoff Length & 12,280 \\
			& Learning Rate & 5e-5 \\
			& Epoch & \{5, 50, 100, 150, 300, 500\} \\
			& Image Max Pixels & 1280 $\times$ 1280 \\
			& Image Min Pixels & 1 $\times$ 1 \\
			\hline
			\multirow{5}{*}{\makecell[l]{Inference}} 
			& Temperature & 0 \\
			& TOP P & - \\
			& TOP K & - \\
			& Num Pre & 12,280 \\
			& Num Ctx & 256,000 \\
			\hline
		\end{tabular}%
	}
	\vspace{-5mm}
\end{table}

To evaluate the performance of MLLMs across defect grading tasks, the accuracy (ACC) and Macro-F1-score (MF1) are adopted as the evaluation metrics. Specifically, Acc and MF1 are calculated using the following Equation \eqref{multi_class_accuracy}-\eqref{macro_f1_score}:

\begin{subequations} \label{eval-metrics-main}
	\begin{align}
		&\text{ACC} = \frac{\sum_{i=1}^{K} N_i}{N_{\text{total}}} = \frac{\text{TP}_1 + \text{TP}_2 + \dots + \text{TP}_K}{\sum_{i=1}^{K} (\text{TP}_i + \text{FP}_i + \text{FN}_i)} \label{multi_class_accuracy} \\
		&\text{MF1} = \frac{1}{K} \sum_{i=1}^{K} \frac{2 \times \text{TP}_i}{2 \times \text{TP}_i + \text{FP}_i + \text{FN}_i} \label{macro_f1_score}
	\end{align}
\end{subequations}
where $K$ is total number of classes in the multi-classification task, $N_i$ is number of correctly classified samples in the $i$-th class, $N_{\text{total}}$ is total number of samples across all classes,  $\text{TP}_i$ is number of true positives for the $i$-th class, $\text{FP}_i$ is number of false positives for the $i$-th class, $\text{FN}_i$ is number of false negatives for the $i$-th class.

\subsection{Rusults of the Commercial MLLMs' Performance}

To maximize the potential of $\mathcal{R}_{dg}$ via ICL and better select the SOTA MLLM to produce fune-tuning dataset Q\&As, we configure four experimental cases in Table \ref{Four prompt settings for testing the commercial MLLMs} below.

%\vspace{-2mm}
\begin{table}[h]
	\centering
	\caption{Four prompt settings for testing the commercial MLLMs}
	\begin{tabular}{@{}lccc@{}}
		\toprule
		\textbf{Case} & \textbf{Only result} & \textbf{DT-based CoT result} & \textbf{Reference images} \\
		\midrule
		Case 1 & $\checkmark$ &  &  \\
		Case 2 &  & $\checkmark$ &  \\
		Case 3 & $\checkmark$ &  & $\checkmark$ \\
		Case 4 &  & $\checkmark$ & $\checkmark$ \\
		\bottomrule
	\end{tabular}
	\label{Four prompt settings for testing the commercial MLLMs}
	\vspace{-2mm}
\end{table}

Specifically, Case 1 contains only pure grading results, Case 2 includes the DT-based CoT results, Case 3 provides the reference images but lacks the DT-based CoT process, and Case 4 not only comprises the DT-based CoT outputs but also incorporates the reference images. The ACC performance of these cases across three tasks is presented in the following Table \ref{Accuracy comparison of different MLLMs in DGPTE}, from which the following conclusions can be drawn:

\begin{table*}[htbp] 
	\centering
	\caption{Accuracy comparison of different MLLMs in DGPTE}
	\resizebox{7in}{!}{
		\footnotesize 
		\begin{tabular}{lcccccccccccc}
			\toprule
			\multirow{2}{*}{\textbf{Model}} & \multicolumn{4}{c}{\textbf{Task 1}} & \multicolumn{4}{c}{\textbf{Task 2}} & \multicolumn{4}{c}{\textbf{Task 3}} \\
			\cmidrule(lr){2-5} \cmidrule(lr){6-9} \cmidrule(lr){10-13}
			& \textbf{Case 1} & \textbf{Case 2} & \textbf{Case 3} & \textbf{Case 4} & \textbf{Case 1} & \textbf{Case 2} & \textbf{Case 3} & \textbf{Case 4} & \textbf{Case 1} & \textbf{Case 2} & \textbf{Case 3} & \textbf{Case 4} \\
			\midrule
			GPT-4.1-2025-04-14 & 53.24\% & 54.64\% & 74.85\% & 89.82\% & 43.49\% & 44.80\% & 46.25\% & 50.66\% & 43.86\% & 48.26\% & 50.50\% & 52.61\% \\
			GPT-4o-2024-11-20 & 54.37\% & 55.61\% & 74.47\% & 88.64\% & 30.18\% & 41.20\% & 46.98\% & 48.03\% & 42.50\% & 39.50\% & 48.95\% & 51.60\% \\
			\textbf{GPT-5-chat} & 51.84\% & 51.24\% & 80.68\% & \textbf{90.26\%} & 30.18\% & 45.72\% & 52.64\% & \textbf{56.58}\% & 42.12\% & 39.50\% & 46.50\% & 45.00\% \\
			Qwen-VL-plus & 53.99\% & 55.81\% & 73.53\% & 84.92\% & 48.64\% & 46.90\% & 50.08\% & 54.56\% & 40.23\% & 42.54\% & 32.86\% & 33.45\% \\
			Qwen3-VL-235B-a22b & 61.18\% & 69.38\% & 81.53\% & 88.22\% & 38.55\% & 52.64\% & 48.64\% & 52.64\% & 43.50\% & 42.65\% & 45.79\% & 48.26\% \\
			\textbf{Qwen3-VL-plus} & 61.80\% & 69.46\% & 75.11\% & 86.05\% & 28.29\% & 34.21\% & 36.78\% & 34.21\% & 45.50\% & 42.74\% & 49.28\% & \textbf{53.80\%} \\
			Gemini-2.5-pro & 53.28\% & 56.37\% & 69.41\% & 82.50\% & 43.49\% & 41.20\% & 46.98\% & 48.93\% & 41.52\% & 43.00\% & 49.86\% & 45.22\% \\
			Hunyuan-t1-vision & 52.96\% & 53.98\% & 78.66\% & 89.03\% & 26.88\% & 28.29\% & 32.92\% & 30.18\% & 39.64\% & 42.66\% & 45.85\% & 47.70\% \\
			Claude-opus-4.5-thinking & 51.44\% & 54.88\% & 71.08\% & 87.59\% & 29.65\% & 30.18\% & 32.92\% & 28.29\% & 44.21\% & 43.68\% & 44.72\% & 45.20\% \\
			Qwen3-VL-8B & 51.68\% & 51.94\% & 56.77\% & 63.06\% & 33.96\% & 27.50\% & 29.88\% & 34.79\% & 22.63\% & 26.16\% & 29.39\% & 30.61\% \\
			\bottomrule
		\end{tabular}
	}
	\label{Accuracy comparison of different MLLMs in DGPTE}
	\vspace{-2mm}
\end{table*}

\begin{itemize}
	\item Gradual improvements in prompt setting can consistently boost the MLLM's ICL capacity. For instance, GPT-5-chat achieves an accuracy of 51.84\% in Case 1 of Task 1, while the accuracy rises to 90.26\% in Case 4 of the same task. Nevertheless, such improvements are subject to a ceiling effect, as exemplified in Task 3 (Case 1: 45.50\%, Case 4: 53.80\%). This indicates that prompt can only yield marginal performance gains when MLLM is inherently incompetent in the DGPTE tasks.

	\item The commercial MLLMs outperform the constrained MLLM (Qwen3-VL-8B) by a significant margin without fine-tuning. In Case 4, GPT-5-Chat attains the highest accuracy in two tasks, reaching 90.26\% for Task 1 and 56.58\% for Task 2, while Qwen3-VL-plus achieves the highest accuracy of 53.80\% for Task 3. By contrast, Qwen3-VL-8B yields only 63.06\%, 34.79\%, and 30.61\%.

	\item The MLLMs' performance exhibits a decreasing trend as the difficulty of the DGPTE task increases. For example, Qwen-VL-Plus achieves accuracies of 84.92\%, 54.56\%, and 33.45\% for Task 1, Task 2, and Task 3, respectively. This indicates that MLLMs can achieve acceptable accuracy on simple tasks but fail to handle challenging ones, even for the commercial MLLMs. This demonstrates the necessity of fine-tuning in this specific field.
\end{itemize}

After obtaining the SOTA MLLM in each DGPTE task (Task 1: GPT-5-chat; Task 2: GPT-5-chat; Task 3: Qwen3-VL-plus), we activate it with $\mathcal{R}^{qa}$ and further generate Q\&As. An Example of Q\&As is provided in Fig. \ref{The details of prompt}.

\subsection{Results Analysis of Proposed Fine-tuning Method for Qwen3-VL-8B}

\begin{figure}[htbp]
	\centering
	\includegraphics[width=3in]{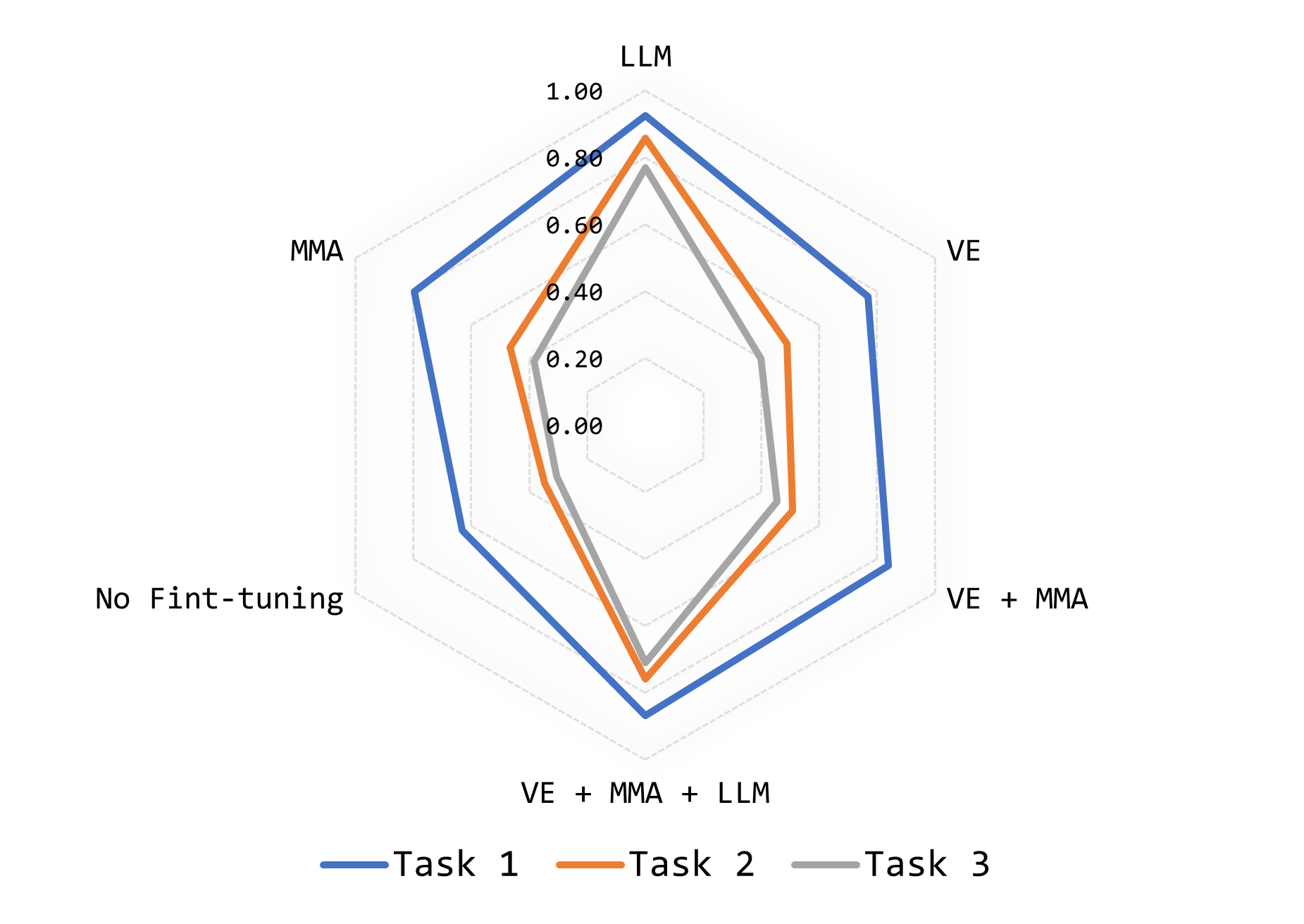}
	\vspace{-6mm}
	\caption{Accuracy comparison of different fine-tuning modules}
	\label{Accuracy comparison of different fine-tuning modules}
	\vspace{-2mm}
\end{figure}

After generating dataset Q\&As, this paper comprehensively fine-tunes different modules of Qwen3-VL-8B to find the optimal ACC scheme, with the result presented in Fig. \ref{Accuracy comparison of different fine-tuning modules}. Specifically, fine-tuning the LLM layer yields SOTA performance with accuracies of 92.63\%, 85.85\%, and 77.04\% across the three tasks. This result significantly outperforms the non-finetuned baseline (63.06\%, 34.79\%, and 30.61\%) and substantially surpasses the SOTA commercial MLLMs in Table \ref{Accuracy comparison of different MLLMs in DGPTE} (90.26\%, 56.68\%, and 53.80\%), which fully validates the feasibility of using a lightweight MLLM to automate tasks in the DGPTE domain.

%\begin{table}[h]
%	\centering
%	\caption{Comparison of Fine-tuning Modules}
%	\begin{tabular}{lcccc}
%		\toprule
%		\textbf{Fine-tuning Module} & \textbf{Task 1} & \textbf{Task 2} & \textbf{Task 3} \\
%		\midrule
%		VE & 76.83\% & 48.82\% & 39.80\% \\
%		MMA & 79.66\% & 46.58\% & 38.27\% \\
%		VE + MMA & 84.01\% & 50.88\% & 45.41\% \\
%		VE + MMA + LLM & 86.75\% & 75.92\% & 70.92\% \\
%		\textbf{LLM} & \textbf{92.63\%} & \textbf{85.85\%} & \textbf{77.04\%} \\
%		No fine-tuning & 63.06\% & 34.79\% & 30.61\% \\
%		\bottomrule
%	\end{tabular}
%	\label{fine_tuning_modules}
%	\vspace{-2mm}
%\end{table}

Our view is that the VE and MMA layers of the MLLM already have the ability to capture key information required for defect grading before fine-tuning, yet the LLM layer lacks the capability to properly retrieve such critical tokens. This phenomenon is attributed to the lack of DGPTE-related query scenarios for MLLMs' pre-training phase. Furthermore, adjusting the VE, MMA, and LLM layers simultaneously yield inferior performance compared to fine-tuning only the LLM layer. This is associated with the scenario of FSL setting, as full-module fine-tuning typically requires a larger sample size; otherwise, it is prone to overfitting.

\begin{figure}[htbp]
	\centering
	\includegraphics[width=3in]{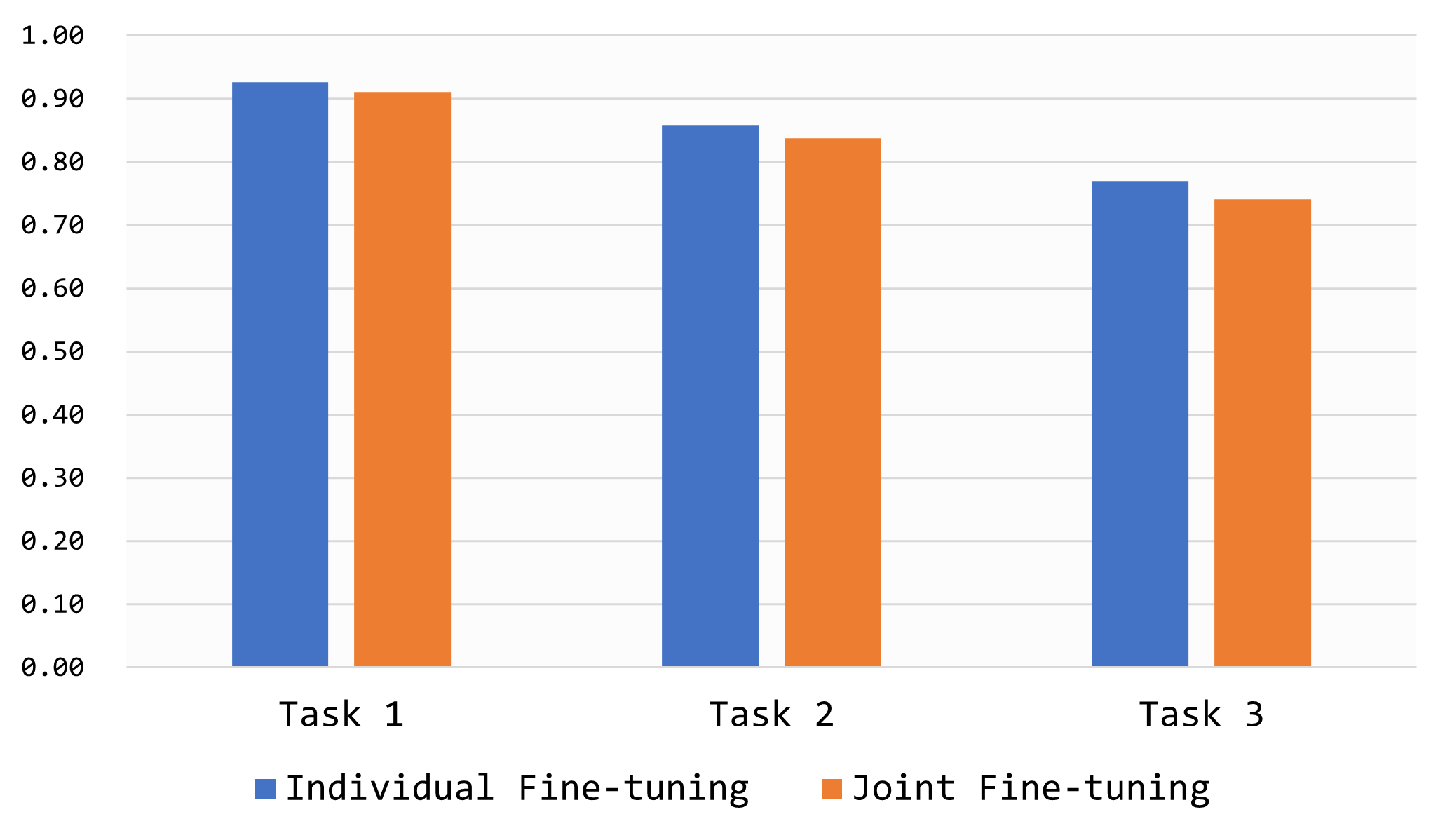}
	\vspace{-6mm}
	\caption{Accuracy comparison of different fine-tuning strategies}
	\label{Accuracy comparison of different fine-tuning strategies}
	\vspace{-2mm}
\end{figure}

Subsequently, multi-task joint fine-tuning is performed on the merged Q\&As from all GDPTE tasks, with the ACC still adopted, and the results are presented in Fig. \ref{Accuracy comparison of different fine-tuning strategies}. Specifically, we find that when small samples from PTEDG tasks are mixed to fine-tune, the model performance experiences a slight decline, yet the drop is within 3\% across all three tasks, and the model still maintains superior performance. This experiment validates the feasibility of using a simple pipeline to simultaneously accomplish multiple industrial visual reasoning tasks.

%\begin{table}[h]
%	\centering
%	\caption{Performance Comparison of Different Fine-tuning Strategies}
%	\setlength{\tabcolsep}{6pt}
%	\begin{tabular}{lccc}
%		\toprule
%		\textbf{Fine-tuning Strategy} & \textbf{Task 1} & \textbf{Task 2} & \textbf{Task 3} \\
%		\midrule
%		Joint Fine-tuning & 91.04\% & 83.79\% & 74.05\% \\
%		Individual Fine-tuning & 92.63\% & 85.85\% & 77.04\% \\
%		\bottomrule
%	\end{tabular}
%	\label{joint_fine_tuning_performance}
%	\vspace{-5mm}
%\end{table}

\begin{figure}[h]
	\centering
	\includegraphics[width=3.5in]{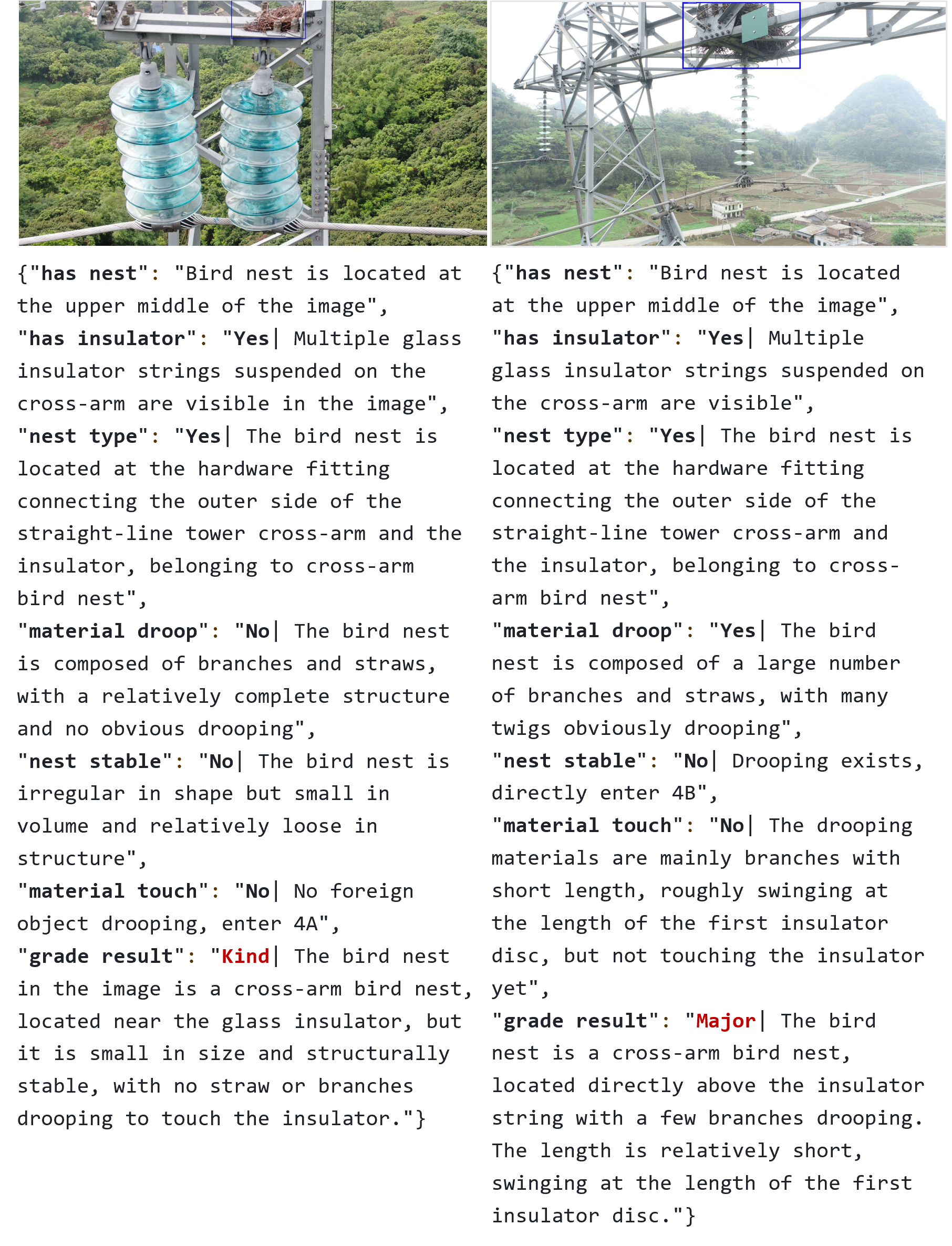}
	\vspace{-6mm}
	\caption{two generated defect grading examples for Task 3}
	\label{The generated defect grading examples for Task 3}
	\vspace{-4mm}
\end{figure}

At the end of this subsection, we present two generated defect grading examples for Task 3 in Fig. \ref{The generated defect grading examples for Task 3}. It is clear that the fine-tuned Qwen3-VL-8B locates visual cues in accordance with expert experience and eventually yields the correct grading result. More  examples are provided in the Appendix.

\subsection{Ablation Experiments}
\textit{1) Impact of Image Position on Performance:} We explore fine-tuning Qwen3-VL-8B with independently organized image and text data fed into it (e.g., all image tokens preceding text tokens). This design is intended to align with the common people interaction habit with MLLM on Graphical User Interface (GUI), where performance is evaluated using ACC and MF1 metrics. The results are presented in Table \ref{image_location_perf}.

\begin{table*}[htbp]
	\centering
	\caption{Performance Comparison by Image Location}
	\resizebox{\textwidth}{!}{
		\begin{tabular}{lcccccccccccc}
			\toprule
			\multirow{3}{*}{\textbf{Image Location}} & \multicolumn{4}{c}{\textbf{Task 1}} & \multicolumn{4}{c}{\textbf{Task 2}} & \multicolumn{4}{c}{\textbf{Task 3}} \\
			\cmidrule(lr){2-5} \cmidrule(lr){6-9} \cmidrule(lr){10-13}
			& \multicolumn{2}{c}{\textbf{Before Fine-tuning}} & \multicolumn{2}{c}{\textbf{After Fine-tuning}} & \multicolumn{2}{c}{\textbf{Before Fine-tuning}} & \multicolumn{2}{c}{\textbf{After Fine-tuning}} & \multicolumn{2}{c}{\textbf{Before Fine-tuning}} & \multicolumn{2}{c}{\textbf{After Fine-tuning}} \\
			\cmidrule(lr){2-3} \cmidrule(lr){4-5} \cmidrule(lr){6-7} \cmidrule(lr){8-9} \cmidrule(lr){10-11} \cmidrule(lr){12-13}
			& \textbf{ACC} & \textbf{MF1} & \textbf{ACC} & \textbf{MF1} & \textbf{ACC} & \textbf{MF1} & \textbf{ACC} & \textbf{MF1} & \textbf{ACC} & \textbf{MF1} & \textbf{ACC} & \textbf{MF1} \\
			\midrule
			At the Front & 42.11\% & 56.83\% & 50.23\% & 53.17\% & 32.80\% & 46.08\% & 34.66\% & 36.52\% & 23.14\% & 38.94\% & 27.50\% & 29.36\% \\
			\textbf{At Corresponding Position} & 59.06\% & 71.03\% & \textbf{92.63\%} & \textbf{92.50\%} & 34.79\% & 49.51\% & \textbf{85.85\%} & \textbf{88.01\%} & 30.61\% & 46.02\% & \textbf{77.04\%} & \textbf{77.32\%} \\
			At the End & 43.86\% & 58.02\% & 50.43\% & 53.30\% & 30.06\% & 44.40\% & 35.33\% & 37.20\% & 25.88\% & 42.12\% & 27.50\% & 30.58\% \\
			\bottomrule
		\end{tabular}
	}
	\label{image_location_perf}
	\vspace{-2mm}
\end{table*}

According to the experimental results, without fine-tuning, placing images at corresponding position in the text yields higher ACC and MF1 scores compared to uniformly placing all images at the front or the end. For example, in Task 2, the ACC and MF1 at the corresponding position reaches 34.79\% and 49.51\%, while those for the other two placement strategies are 32.80\% and 46.08\% (at the beginning), 30.06\% and 35.33\% (at the end). After fine-tuning, placing images at the corresponding position leads to breakthrough improvements in both ACC and MF1. However, when all images are placed at the front or end, the ACC is improved whereas the MF1 shows a reverse decline. For instance, in Task 1, the ACC increases from 42.11\% before fine-tuning to 50.24\% after fine-tuning, while the MF1 decreases from 56.83\% to 53.17\% (at the front). We analyze that although the model accuracy is slightly improved after fine-tuning, this is not indicative of genuine performance enhancement; instead, the model loses its discriminative capability and merely makes blind category predictions. We think that this phenomenon is associated with the fact that MLLM processes mixed image-text data at their corresponding positions during the pre-training phase.

\textit{2) Impact of Image Resolution on Performance and Computational Consumption:} Computational consumption and single-image inference latency of Qwen3-VL-8B during both training and inference phases is the core concern. This paper validate the relevant results by resampling images to different sizes, with the results showed in Table \ref{size_overhead_mf1}. Based on the results, image resolution exerts a crucial impact on model performance: higher image clarity contributes to improved task accuracy (e.g., the average accuracy across the three tasks increased from 68.55\% at a resolution of 224 to 85.17\% at 1280). However, this is accompanied by a significant increase in training memory footprint and training time. For instance, fine-tuning the model is no longer feasible with 640 GB of memory when the image resolution is set to 2560, while the training time reaches 9.469 hours at a resolution of 1280. Similarly, the memory and time required for model inference also rise with an increase in image resolution (at 1280, the inference time is approximately 9 seconds per image with a memory requirement of around 42 GB). 

Given the timeliness requirements of industrial applications, a trade-off must be made between these factors, and this study identifies 1280 as the optimal resolution.

\begin{table*}[htbp]
	\centering
	\caption{Performance Overhead by Different Input Sizes}
	\begin{tabular}{lcccccc}
		\toprule
		\textbf{Size} & \textbf{ACC} & \textbf{MF1} & \textbf{Training Time (h)} & \textbf{Training VRAM (GB)} & \textbf{Inference Time (s)} & \textbf{Inference VRAM (GB)} \\
		\midrule
		p224  & 68.55\% & 67.76\% & 2.246   & 255.996  & 5.046  & 12.969  \\
		p448  & 72.76\% & 73.88\% & 2.956   & 282.395  & 5.140   & 22.495  \\
		p768  & 77.42\% & 80.02\% & 3.907   & 325.293  & 6.975  & 31.850  \\
		\textbf{p1280} & \textbf{85.17\%} & \textbf{85.94\%} & 9.469   & 489.999  & 9.236  & 42.285  \\
		p2560 & - & - & -       & -        & 10.564 & 74.592  \\
		\bottomrule
	\end{tabular}
	\label{size_overhead_mf1}
	\vspace{-5mm}
\end{table*}

%\textit{3) Impact of Fine-tuning Epochs on Performance:}
%In the final ablation experiment, we investigate the impact of different fine-tuning epochs on model performance. The prevalent research domain of multimodal computer vision emphasizes the importance of minimizing the number of fine-tuning epochs as much as possible. To verify the effectiveness of this principle for industrial visual reasoning tasks such as PTEDG, the relevant experimental results of this study are presented in Table \ref{acc_epoch}.

%\begin{table}[h]
%	\centering
%	\caption{Relationship between Epoch, ACC, MF1 and Training Time}
%	\setlength{\tabcolsep}{6pt}
%	\begin{tabular}{lccc}  
%		\toprule
%		\textbf{Epoch} & \textbf{ACC} & \textbf{MF1} & \textbf{Training Time (s)} \\  
%		\midrule
%		2   & - & - & - \\  
%		10  & - & - & - \\
%		50  & - & - & - \\
%		150 & - & - & - \\
%		300 & - & - & - \\
%		400 & - & - & - \\
%		\bottomrule
%	\end{tabular}
%	\label{acc_epoch}
%	\vspace{-5mm}
%\end{table}

\vspace{-2mm}

\section{Conclusion}\label{Five}
Due to a series of challenges, such as severe sample imbalance, reliance on human knowledge and the need for technical visual reasoning, the refined defect grading of power transmission equipment tasks haven't fully been automated. To bridge this gap, this paper proposes an MLLM-enabled cost-effective grading framework. After strategically generating the interpretable grading Q\&As, the lightweight Qwen3-VL-8B is supervised fine-tuned based on a small number of this set. Extensive experiments on three real-world DGPTE tasks verify the fine-tuned MLLM achieving accuracy far surpassing that of the best commercial MLLMs and comparable to manual grading in target tasks. This framework still maintains good performance when extended to multi-task joint fine-tuning, which reveals the potential of a single and offline-deployable MLLM to handle multiple DGPTE tasks simultaneously. 

To further enhance engineering practicality, future work will focus on fine-tuning a lightweight MLLM capable of multi-turn interaction to meet the different personalized grading needs. In addition, the solution of training a professional world MLLM to fully understand the power industry and improve grading accuracy remains to be further explored.

\addcontentsline{toc}{chapter}{Bibliography}
\ifCLASSOPTIONcaptionsoff
\newpage
\fi 
\bibliographystyle{IEEEtran}
\bibliography{IEEEabrv,paper}

\end{document}